\documentclass{svproc}

\usepackage[nolist]{acronym}
\usepackage{bm}
\usepackage{booktabs}
\usepackage{color}
\usepackage{cite}
\usepackage{graphicx}
\usepackage{multirow}
\usepackage{tabularx}
\usepackage{siunitx}

\usepackage[normalem]{ulem}

\usepackage{enumitem}
\usepackage{url}
\usepackage{hyperref} 

\newacro{hri}[HRI]{Human-Robot Interaction}

\begin{document}
\mainmatter              
\title{Hearing the Robot's Mind: Sonification for Explicit Feedback in Human-Robot Interaction}
\titlerunning{Hearing the Robot's Mind}  
\author{Simone Arreghini\inst{} \and Antonio Paolillo \inst{}\and Gabriele Abbate \inst{} \and
Alessandro Giusti \inst{} }
\authorrunning{Simone Arreghini et al.} 
\tocauthor{Simone Arreghini, Antonio Paolillo, Gabriele Abbate, Alessandro Giusti}
\institute{Dalle Molle Institute for Artificial Intelligence (IDSIA), USI-SUPSI, \\Lugano, Switzerland 
\\ \email{name.surname@idsia.ch}}
\maketitle              

\begin{abstract}
Social robots are required not only to understand human intentions but also to effectively communicate their intentions or own internal states to users. 
This study explores the use of sonification to provide explicit auditory feedback, enhancing mutual understanding in HRI.
We introduce a novel sonification approach that conveys the robot's internal state, linked to its perception of nearby individuals and their interaction intentions.
The approach is evaluated through a two-fold user study: an online video-based survey with $26$ participants and live experiments with $10$ participants. 
Results indicate that while sonification improves the robot's expressivity and communication effectiveness, the design of the auditory feedback needs refinement to enhance user experience.
Participants found the auditory cues useful but described the sounds as uninteresting and unpleasant. 
These findings underscore the importance of carefully designed auditory feedback in developing more effective and engaging \ac{hri} systems.
\end{abstract}
\section{Introduction}\label{sec:intro}

Many novel applications of robotics envision close interaction with humans in everyday life settings, both in private~\cite{cui2020mmpd} and public spaces such as hospitals~\cite{Gonzalez:as:2021}, hotels~\cite{Choi:jhmm:2020} or museums~\cite{hellou2022technical}.
For effective interactions in such scenarios, robots must be capable of operating gently, and responding to different human needs and behaviors; in practice, they need emotional~\cite{sirithunge2019proactive} or social intelligence~\cite{Zaraki:icrm:2014,nocentini2019survey}. 
This new class of \emph{social} robots is rapidly growing, setting new technological and research challenges.
In fact, in such challenging scenarios, robots have to accurately detect and interpret human intentions and behaviors.
At the same time, the robot's awareness of the human's intention must be made clear to the users, to promote smooth and efficient interaction.

\begin{figure}[t]
    \centering
    \frame{\includegraphics[trim={0.0cm 0.0cm 0.0cm 0.0cm},clip,width=0.485\columnwidth]{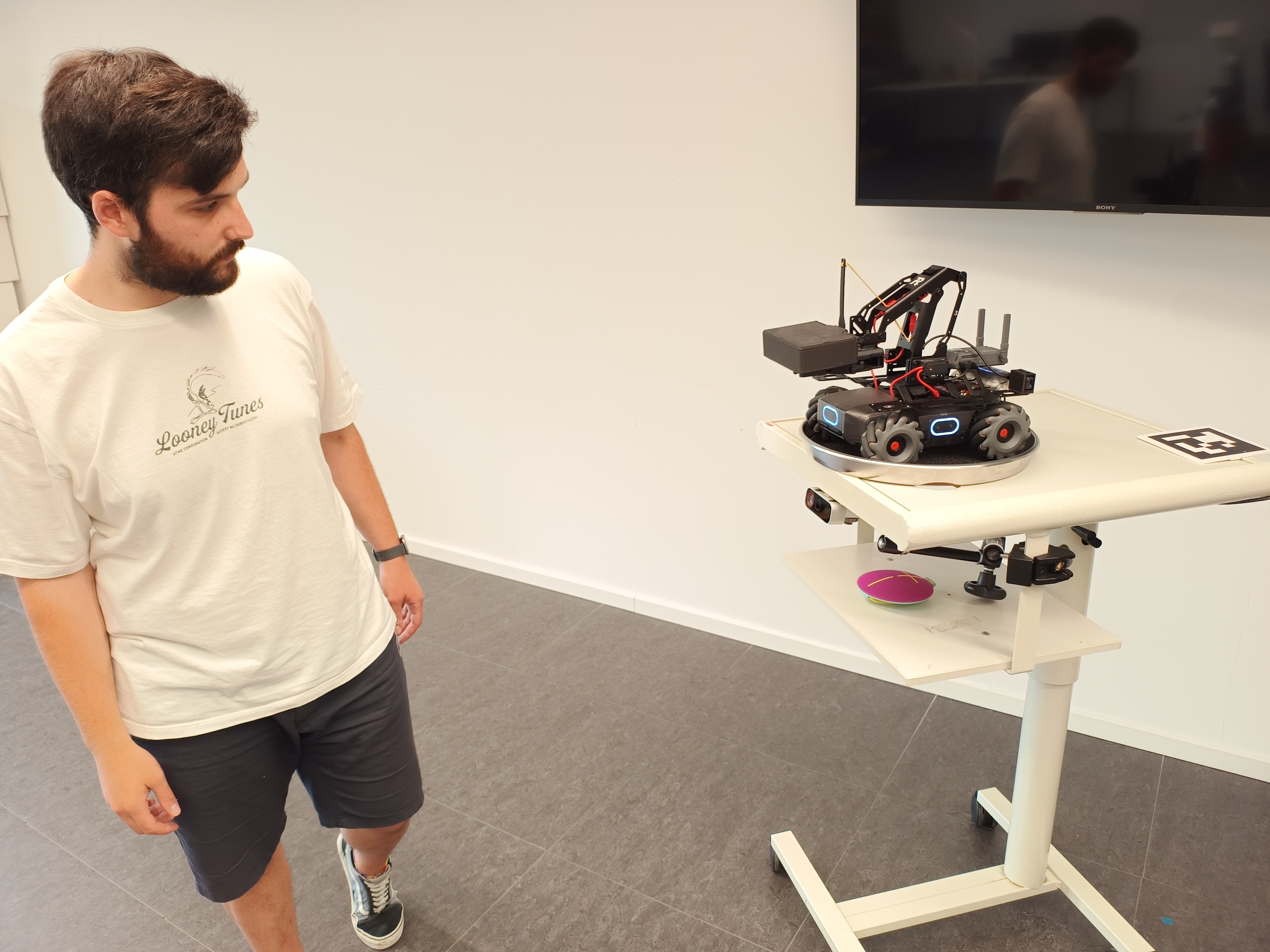}}%
    \hspace{0.02\columnwidth}%
    \frame{\includegraphics[trim={0.0cm 0.0cm 0.0cm 0.0cm},clip,width=0.485\columnwidth]{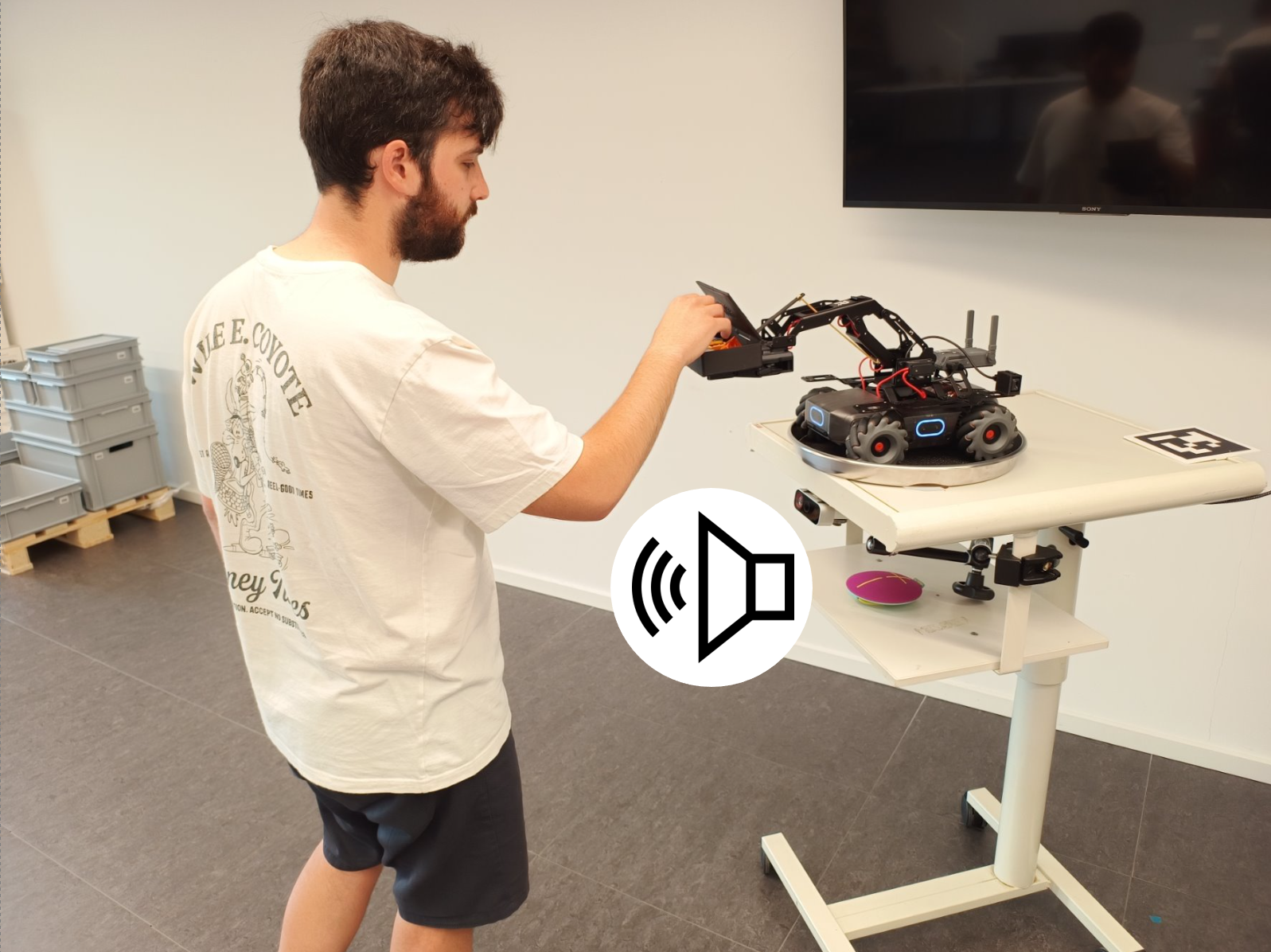}}%
    \caption{A robot offering chocolate treats to a visitor of a public building uses audio cues to inform nearby people of its internal state.}%
    \label{fig:interaction}
\end{figure}

Consider a robot tasked with offering chocolate treats to the visitors of a public building, as illustrated in Fig.~\ref{fig:interaction}. 
To ensure the positive involvement of all the users, even those skeptical or shy, the robot must be capable of exteroceptive perception, detecting nearby individuals, predicting their intentions, and promptly reacting to those who show interest. 
During this process, it is also desirable that robots express their internal state and beliefs about the world to the users using non-verbal communication channels; which may, in some cases, ``even say some things with greater facility and efficiency than with words''\cite{mehrabian2017nonverbal}.
Technically, detecting users who enter the robot's social space is feasible with current state-of-the-art sensors. 
Building on these technologies, our previous work~\cite{Abbate:ras:2024,Arreghini:icra:2024} developed a perception pipeline to discern users' intentions to interact with the robot. 
These self-supervised learning methods utilize body pose and motion cues to predict the likelihood of future interactions~\cite{Abbate:ras:2024}. 
Integrating facial features and a mutual gaze detector (specifically designed for \ac{hri} applications~\cite{Arreghini:hri:2024}) further enhances prediction performance~\cite{Arreghini:icra:2024}. 
Once user intention is predicted, the robot can use such information to trigger appropriate robot behaviors. 
This approach has been thoroughly tested in both controlled and uncontrolled scenarios, demonstrating the robot's ability to perceive the user intention~\cite{Arreghini:iros:2024} -- a form of indirect, nonverbal communication from the user to the robot.

In this work, we aim to establish a two-way communication stream by exploring how sonification techniques~\cite{walker2011theory} can facilitate the flow of information from the robot to the users and examine how users perceive these sounds.
In assessing these assumptions, we propose the following contributions.
Firstly, we present a sonification approach to convey the robot's internal state, linked to its perception of the surroundings, and more specifically to the estimated probability of  interaction of nearby individuals.
Secondly, we evaluate how the sonification of the robot's internal state is perceived by people, through an ad-hoc questionnaire.    
The remainder of the paper is structured as follows.
Section~\ref{sec:related_work} reviews the state-of-the-art and Sec.~\ref{sec:experimental_setup} details the experimental setup used to deploy our robot.
The results 
are presented in Sec.~\ref{sec:results}, whereas final discussions and conclusions are in Sec.~\ref{sec:conclusions}.
\section{Related Work}\label{sec:related_work}
In \ac{hri}, social robots are designed to interact as peers with the human counterpart~\cite{goodrich2008human}.
However, robots that are left to operate autonomously in the direct presence of human beings face a multitude of challenges that vary depending on the specific application.
In interactions between humans, non-verbal implicit or explicit behaviors are of utmost importance, as they convey emotions and attitudes, and enhance verbal communication~\cite{mehrabian2017nonverbal}.
In \ac{hri} scenarios this kind of communication can be divided into two separate channels depending on the direction of the information flow: human-to-robot and robot-to-human.
In real social contexts, understanding human needs is often linked to interpreting human intentions, with non-verbal cues playing a crucial role~\cite{Gasteiger:ijsr:2021}.
This topic has been extensively studied across various \ac{hri} contexts: navigation~\cite{Agand:icra:2022}, collaborative tasks~\cite{Belardinelli:iros:2022,Vinanzi:icdler:2019}, and interpreting social behaviors~\cite{Zaraki:icrm:2014,Gaschler:iros:2012}, for example.
Detecting user intentions to interact, such as through body posture alone, is particularly useful for early-stage \ac{hri} reactions~\cite{Abbate:ras:2024}.
Human intention detection is further enhanced by incorporating gaze cues~\cite{Arreghini:icra:2024}, which are widely recognized as strong indicators of user intentions~\cite{belardinelli:thri:2023}.

However, non-verbal communication between humans and robots also flows in the opposite direction, with robots conveying their intentions to human counterparts~\cite{Saunderson:ijsr:2019}.
Robots can leverage multiple sources of non-verbal communication to provide different degrees of feedback to the user. 
For instance, they can simply use their spatial positioning applying proxemics concepts~\cite{Rios:ijsr:2015} to convey their navigation goal or motion intentions~\cite{Marquardt:pc:2012}.
They might mimic expressive gestures and body language from humans to communicate clues about their internal state~\cite{venture2019robot} or achieve even more engaging visual feedback through the use of lights and LEDs~\cite{haas2017exploring}.
Another way for enhancing feedback toward humans is to accompany motions with sounds using sonification techniques to increase users' awareness about the robots' status and movements around people~\cite{zahray2020robot}.
Sonification is the process of converting data into non-speech audio signals~\cite{walker2011theory}. This auditory display technique allows people to interpret and understand data through sound, often in situations where visual representation might be impractical or less effective.
In the robotics domain, using sonification approaches for non-verbal communication in \ac{hri} scenarios has already been explored in many works~\cite{zhang2023nonverbal}, with efforts done even towards a unified approach for sound generation~\cite{zhang2022sonifyit} for general robotic platforms.
Robotic agents can use these means of feedback to provide the user with information about their motion intents~\cite{pascher2023communicate}, check and monitor engagement~\cite{maniscalco2022bidirectional} 
or express emotions~\cite{venture2019robot} enabling more effective, efficient, and well-perceived human-robot interactions.
In this work, we integrate a sonification approach with our pipeline capable of estimating the intentions of individuals interacting with the robot. 
Our objective is to generate auditory feedback from the robot that communicates an internal state influenced by the user's behavior itself.
\section{Experimental Setup}\label{sec:experimental_setup}

\subsection{Robotic system}
\begin{figure}[!t]
\centering
 \frame{\includegraphics[trim={4.0cm 2.0cm 4.0cm 3.0cm},clip,width=0.75\textwidth]{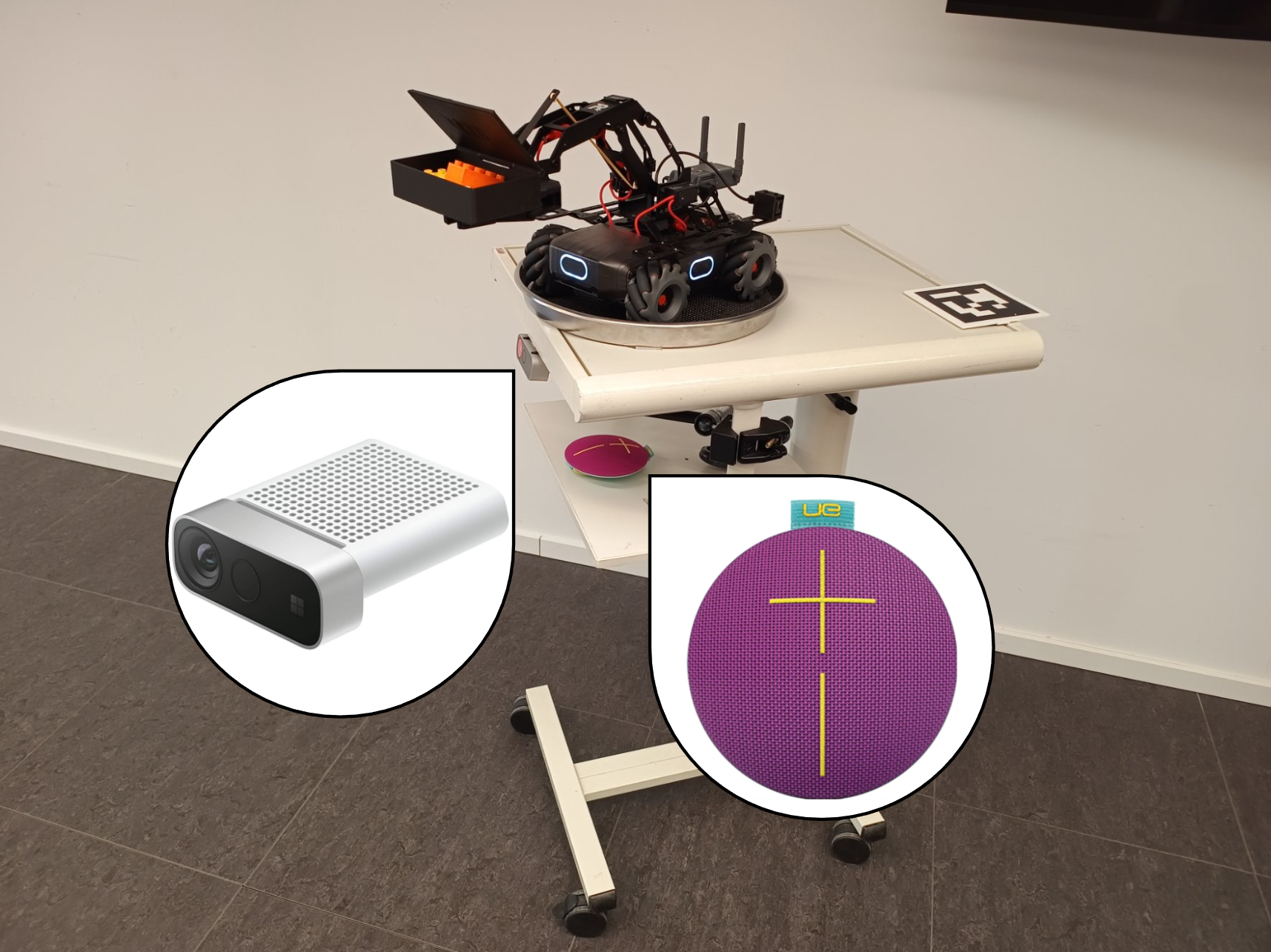}}
\caption{Experimental setup with a zoomed view of the sensor and the speaker used to elaborate our two-way communication between a robot and a user.}
\label{fig:experimental_setup}
\end{figure}

The setup used in our tests is illustrated in Fig.~\ref{fig:experimental_setup}.
We use a DJI RoboMaster EP Core\footnote{https://www.dji.com/ch/robomaster-ep-core}, a compact mobile robot with omnidirectional wheels, LED lights, and an arm with $2$ degrees of freedom. 
The arm is endowed with a $1$ degree of freedom gripper and has a
payload of \SI{0.3}{kg}. 
This is enough to firmly hold a small box with a lid, filled with chocolate threats, which automatically opens when the arm extends. 
The robot is positioned on a table \SI{1.10}{m} high, with a border that prevents accidental falls. 
For sensing requirements, we use an Azure Kinect\footnote{https://learn.microsoft.com/en-us/azure/kinect-dk/system-requirements} mounted just below the table surface to closely match the robot's perspective.
This RGB-D camera can stream $4$K images and track and extract information about people in the robots'  field of view thanks to the built-in human body tracking capability.
From the images provided by this sensor, we extract facial landmarks using the MediaPipe\footnote{\url{https://ai.google.dev/edge/mediapipe/solutions/vision/face_landmarker}} 
Python package. 

Facial data and body tracking information are then fed to an interaction intention classifier~\cite{Arreghini:icra:2024};
%
its output, for each tracked user, represents the predicted probability that the user will interact with the robot.  It is close to $0$ for users whose behavior and body language imply that they are not willing to interact and likely won't engage with the robot; it rises to $1.0$ as a user looks at the robot, and approaches it.  This value, which is computed at each frame, is smoothed over time with an exponential moving average, with a time constant equal to $1$ second; we represent this smoothed probability value with variable $p \in [0,1]$. 

The robot communicates with a user in several ways: moving its body and arm; lighting and changing color to its LEDs; and generating sound through a speaker.
In idle conditions the robot stays still, with the body aligned to the sensor's forward direction; its LEDs are weakly lit in white -- to indicate that the robot is turned on -- and the arm is fully retracted with the box lid closed.

The robot motion reaction is triggered once the predicted interaction intention probability exceeds $p > 0.85$. This value allows for maximizing the precision and recall of the perception pipeline.
In this instant, the robot rotates in place to orient the arm towards the selected person, lights up its LEDs, and extends its arm forward. The arm movement causes the lid of the box held by the robot to open automatically, allowing the person to access the chocolate treats.
During this activation phase, the LEDs are lit with an intensity proportional to the predicted probability, varying the color from blue to yellow, from low to high probability values.  
The robot retracts its arm and reorients itself to the neutral direction either when the user picks a chocolate treat or once $p$ falls below a lower threshold $p < 0.75$ for at least $1$ second.  This simple hysteresis scheme ensures that the robot does not retract its arm too soon after extending it.
\subsection{Sound generation}

\begin{figure}[!t]
{\includegraphics[width=\textwidth]{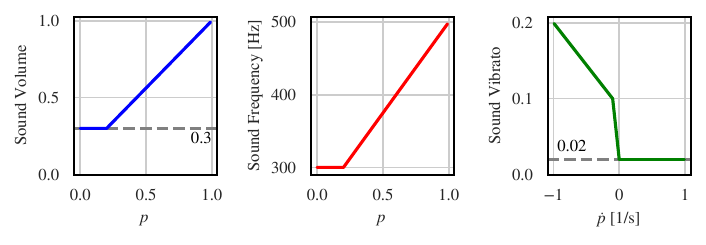}}
\caption{Plots of the piecewise linear transfer functions used to calculate the sound parameters: volume and pitch depend on $p$, whereas the amount of vibrato effect depends on its rate of change. 
}
\label{fig:prob_to_sound}
\end{figure}
Sound is synthesized in real-time by a simple signal processing pipeline implemented through the Pyo\footnote{\url{https://pypi.org/project/pyo/}} Python package.  The pipeline generates sound based on three parameters that are updated after each frame is processed: \emph{volume}, \emph{frequency} (i.e. pitch), and \emph{vibrato}.

As illustrated in Fig.~\ref{fig:prob_to_sound}, volume and frequency depend on the current value of $p$. The sound becomes increasingly loud and high-pitched as the probability rises, which is designed to convey a sense of \emph{excitement} of the robot as it becomes increasingly confident that the user is going to interact.

Below the $p=0.2$ threshold, both volume and frequency bottom out to a low constant value, causing the robot to output a weak but audible low-pitched sound whenever a person is perceived in front of the robot, even when not interested in interacting.  In contrast, when no person is perceived, the robot emits no sound at all. 
This feature is designed to convey the robot's \emph{awareness} of the user's presence, audibly differentiating the case in which no user is tracked from the case in which at least one user is seen.

\begin{figure}[!t]
{\includegraphics[width=\textwidth]{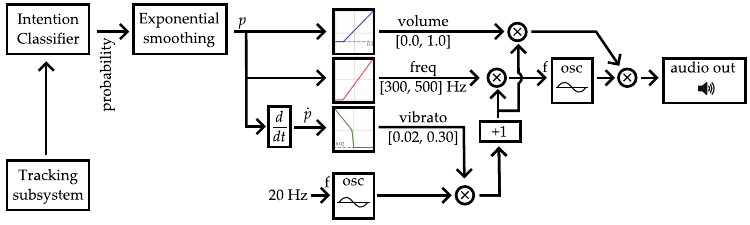}}
\caption{Diagram of the sound synthesis pipeline}
\label{fig:pipeline}
\end{figure}

Lastly, the amount of vibrato effect in the generated sound is determined by the rate of change of $p$; when $p$ is not decreasing, vibrato has a constant, low value of $0.02$: this yields an almost-pure sound with a barely noticeable rapid fluctuation, which gives it a little more interest.  In contrast, when $p$ decreases (indicating that the user is indeed not as interested in the offer as the robot previously thought), the robot expresses \emph{frustration} through a noticeable increase in vibrato.  Then, the sound pitch and volume rapidly fluctuate at a rate of 20 Hz, giving the sound a peculiar character that one might associate with disappointment, sadness, or frustration.  The sound synthesis pipeline is summarized in Fig.~\ref{fig:pipeline}

Once a tracked individual takes the chocolate, the robot detects the event\footnote{It would be nice, at this stage, to emit a brief squeal of delight, but this is not currently implemented.}, retracts its arm, ceases emitting sound, and stops reacting to that specific user.
Due to privacy issues, the user differentiation doesn't use any facial recognition software but is based on a simple body tracking algorithm.  Therefore, if a person who has previously interacted with the robot exits its sensor field of view and then enters again, it will be treated as a new user, with the robot again reacting to its presence.  

\subsection{Evaluation}

\begin{figure}[t!]
    \centering
    \begin{minipage}[t]{0.32\textwidth}
        \includegraphics[width=\textwidth]{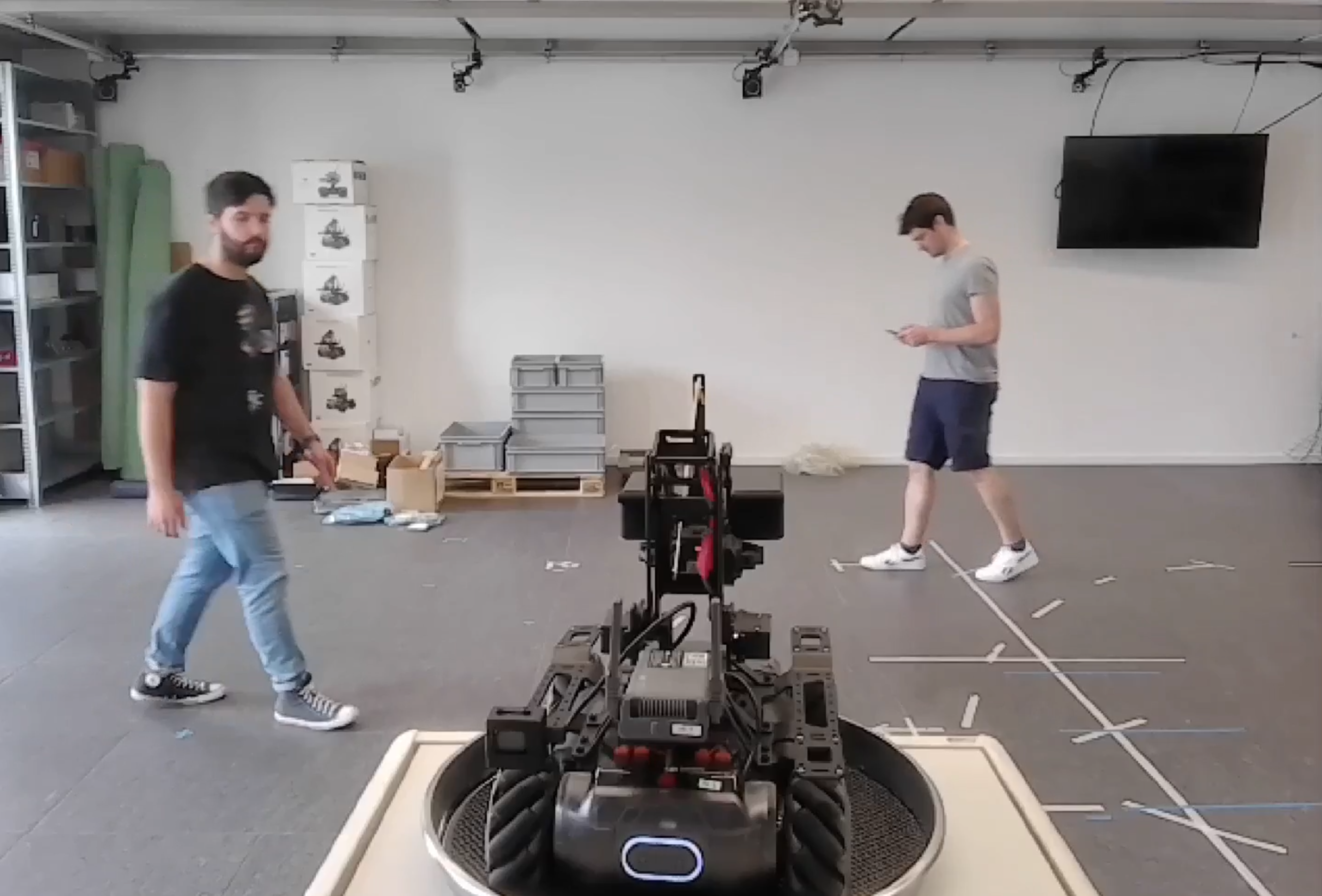}
    \end{minipage}
    \hfill
    \begin{minipage}[t]{0.32\textwidth}
        \includegraphics[width=\textwidth]{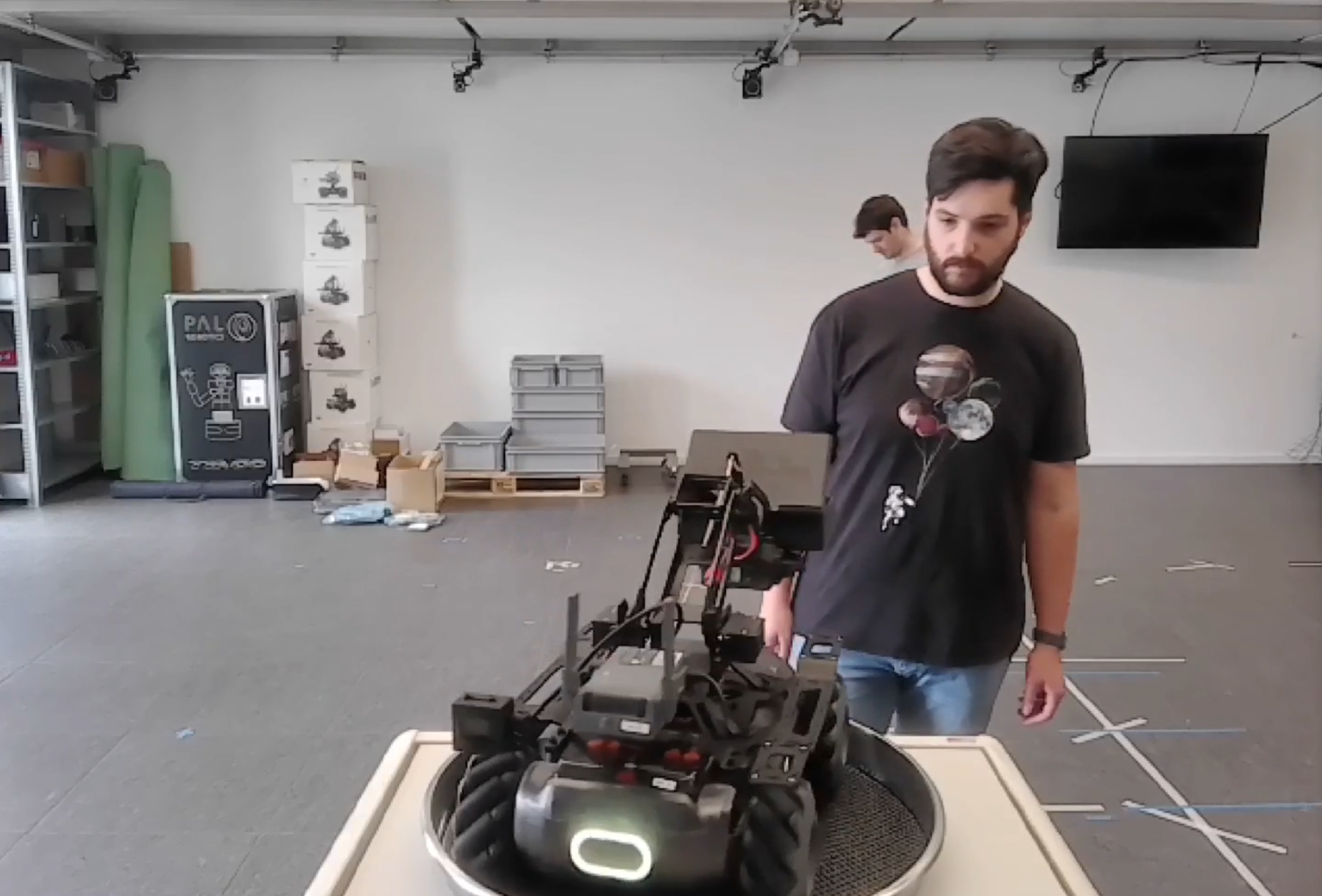}
    \end{minipage}
    \hfill
    \begin{minipage}[t]{0.32\textwidth}
        \includegraphics[width=\textwidth]{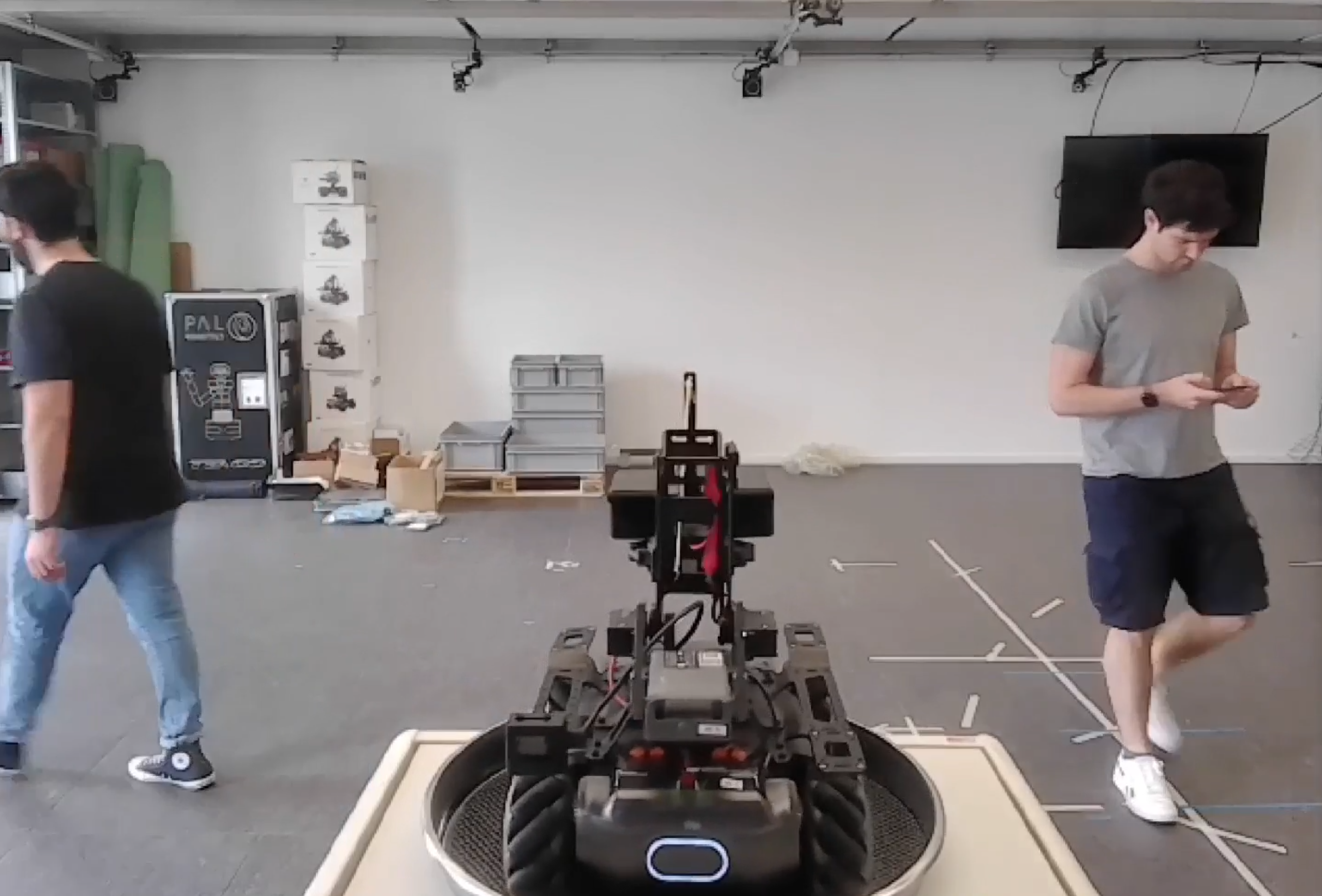}
    \end{minipage}

    \begin{minipage}[t]{\textwidth}
        \centering
        {(a) Approaching} \hspace{2.0cm} {(b) Offering} \hspace{2.0cm} {(c) Distancing}
    \end{minipage}
    
    \begin{minipage}[t]{\textwidth}
        \includegraphics[width=\textwidth]{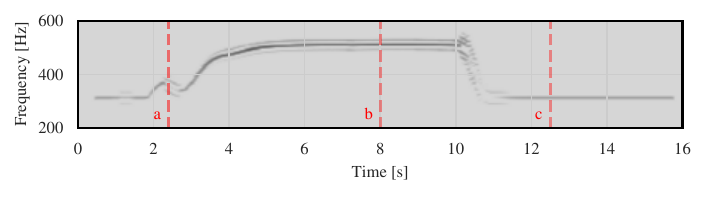}
    \end{minipage}
    \caption{Spectrogram of the sound feedback during an interaction sequence in which one user approaches the robot looking very interested (a), the robot keeps offering them the chocolate for a long time (b, from second 4 to second 10), then the user suddenly leaves without actually taking the chocolate (c, second 10 to 12).  This sequence corresponds to timestamps 0:39 to 0:55 of the attached video~\url{https://youtu.be/Cn9dQBznWzY} }
    \label{fig:snapshots}
\end{figure}

Our approach is evaluated with a user study carried out in three separate sessions: an online video-based survey, and two live robotic experiments. 
In the video-based survey, participants are shown a brief video containing some interaction sequences between the robot and two actors.
A video with interaction sequences very similar to the online survey, with an added top-view map, is available here:~\url{https://youtu.be/Cn9dQBznWzY}.
Some snapshots of the video are presented in Fig.~\ref{fig:snapshots} together with a spectrogram representing the sound during the interaction.

In the live experiments, participants are left to interact with the robot in two tests of around $1$ minute each. 
During this time they experience both the same sonification approach captured in the video, but also a baseline without any auditory feedback whatsoever. 
To prevent any learning effect from happening, the order in which each condition is presented to the users is alternated between subsequent participants.
After all sessions, the participants are asked to fill out a questionnaire composed of $10$ points probing their perceptions about the general interaction (in all conditions) and the sound feedback (where applicable).
The scores for each point are reported in Tab.~\ref{tab:questions}.
All the answers, given using a $5$-point scale, have been standardized to a higher-is-better scale, with $1$ being the lowest value and $5$ the highest. 
In this way, the higher the average score for each point, the better it is from a user experience point of view. 
However, the description of the highest and lowest score values may vary between the different questions.
The first three questions are designed to probe the general impressions of the robot's behavior. 
Users can rate the robot behavior from ``Not at all expressive'' (1) to ``Very expressive'' (5) in P1; how well they understood the robot behavior from ``Not at all'' (1) to ``Completely'' (5) in P2; and how effective was the robot in communicating its state from ``Not effective'' (1) to ``Very effective'' (5) in P3.
Point $4$ is designed to probe the user perceptions about the sound feedback as a whole where users rate the sound feedback impact on the interaction with a range from ``Very negative'' (1) to ``Very positive'' (5).
While these last questions have been created ad-hoc for our experiments, points from P5 to P10 are taken with little to no adaptation from the BUZZ scale~\cite{tomlinson2018buzz}, which is commonly used to rate auditory stimulation in \ac{hri}.
In this scale, users are asked to rate the degree to which they agree with a particular statement using a score from $1$ (``strongly disagree'') to $5$ (``strongly agree'').
\begin{table*}[!tb]
    \centering
    \caption{Different questionnaire points submitted to the participants of our user study, with the average and standard deviation of the corresponding scores.}
    \begin{tabular}{clcccc} 
        \toprule
        \multicolumn{2}{l}{\textsc{Points}} & \multicolumn{3}{c}{\textsc{Scores}} \\
         &  & \multicolumn{1}{c}{\textsc{Video-based}} &  \multicolumn{2}{c}{\textsc{Live Experiments}}\\
        \# & Text & \multicolumn{1}{c}{\textsc{Survey}} & Sonification & Baseline \\
        \midrule
        \multirow{2}{*}{P1} & How expressive did you find the robot & \multirow{2}{*}{$3.7 \pm 0.9$} & 
        \multirow{2}{*}{$4.3 \pm 0.5$} & \multirow{2}{*}{$3.4 \pm 0.7$}\\
         & in this video? & & & \\
        \multirow{2}{*}{P2} & How well did you understand the robot's & \multirow{2}{*}{$4.1 \pm 1.0$} &   \multirow{2}{*}{$4.5 \pm 0.5$} & \multirow{2}{*}{$3.8 \pm 0.8$}\\
         & behavior during the interactions? & & & \\
        \multirow{2}{*}{P3}  & How effective was the robot in & \multirow{2}{*}{$4.0 \pm 1.0$} &           \multirow{2}{*}{$4.5 \pm 0.5$} & \multirow{2}{*}{$3.7 \pm 0.8$}\\
         & communicating its state? & & & \\
        \multirow{2}{*}{P4} & How did the sound feedback affect your & \multirow{2}{*}{$4.0 \pm 1.2$} &    \multirow{2}{*}{$4.3 \pm 0.7$} & \multirow{2}{*}{-}\\
         & perception of the robot's behavior? & & & \\
        P5  & The sounds were helpful & $4.3 \pm 0.6$ & $4.0 \pm 0.9$ & -\\
        P6  & The sounds were interesting & $3.1 \pm 1.1$ & $3.1 \pm 1.0$ & -\\
        P7  & The sounds were pleasant & $2.2 \pm 1.3$ & $2.4 \pm 1.0$ & -\\
        P8  & It was confusing to listen to these sounds & $4.1 \pm 1.1$ & $4.1 \pm 0.6$ & -\\
        P9  & The sounds were easy to understand & $4.1 \pm 0.8$ & $3.8 \pm 0.8$ & -\\
        P10 & The sounds were relatable to the robot state & $4.3 \pm 0.6$ & $4.1 \pm 0.9$ & -\\
        \bottomrule
    \end{tabular}
    \label{tab:questions}
\end{table*}

In all cases, at the end of the questionnaire, there is an \emph{Additional Comments} section which is used to gather possible remarks not covered by the different questions or improvements to the overall pipeline. 
\section{Results}\label{sec:results}
\subsection{Video-based survey study}

A total of $26$ people ($61.5\%$ M, $38.5\%$ F) completed the online video questionnaire; the demographics is as follows: $61.5\%$ of the participants are between $20$ and $29$ years old; $15.4\%$ in the range $30-39$; $11.5\%$ in $40-49$; $7.7\%$ in $50-59$; and $3.8\%$ are more the $60$ years old.
Experience with robots, rated on a scale from 1 to 5, reports most ratings between 2 and 4, with both genders having similar median ratings despite the broader distribution among males. Indicating on average a moderate level of experience in dealing with robots of the population.

\begin{figure}[!t]
{\includegraphics[width=\textwidth]{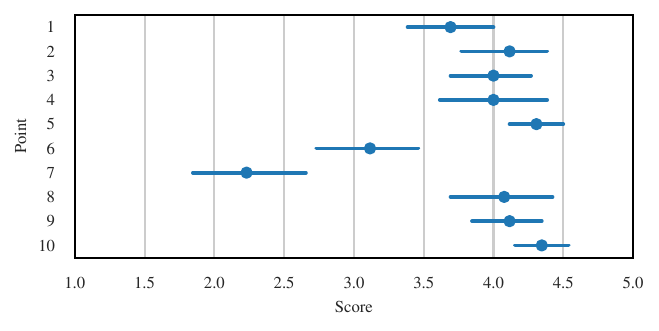}}
\caption{Point plots of the answers to the Video-based online survey. The points represent the mean value, whereas the line is the $90\%$ confidence interval.}
\label{fig:single_questions_video}
\end{figure}

Table~\ref{tab:questions} reports the values of the scoring average and standard deviation of the questionnaire answers in all conditions. 
However, a more visual representation of the answers to the video-based survey is provided in Fig.~\ref{fig:single_questions_video}.
Using this point plot we can appreciate how the majority of the questions' average scores are around $4.0$ or above indicating a satisfactory user experience. 
Overall, participants found the robot to be moderately expressive (P1) reporting a good understanding of its behavior (P2) and deeming it effective in communicating the robot's internal state (P3).
Regarding the sonification specifically, participants generally found the sounds to positively affect the interaction (P4). 
The auditory feedback is deemed helpful (P5) and easy to understand (P9), generally perceiving the sounds as not confusing (P8) as clearly linked to the robot's state (P10). 
However, opinions greatly vary on the subjective qualities of the sounds: while some found them marginally interesting ($P6$), the majority perceived the sounds as not pleasant ($P7$) indicating a possible need for refinement in the auditory design to enhance overall user satisfaction and engagement.
The trends explained above are further validated by some of the additional comments left at the end of the questionnaire.
Some participants repeated the helpfulness of the sound feedback in helping the communication of the robot's internal state. However, the same people remarked how the high-pitched sound is perceived as very annoying, with some describing it as ``an alarm'' or ``a cornered animal that would scare them away instead of attracting users to the robot''.
\subsection{Live experiments}
\begin{figure}[!t]
{\includegraphics[width=\textwidth]{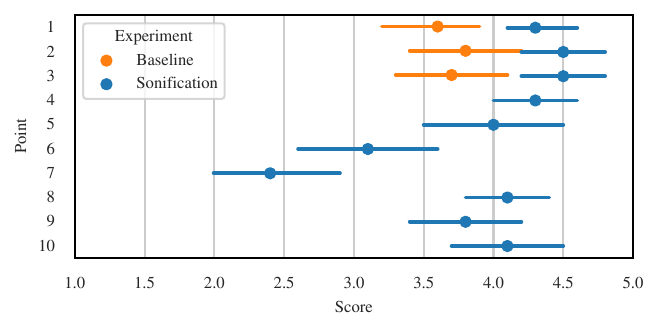}}
\caption{Point plots of the answers to the questionnaire for the live experiments. In orange are the scores for the Baseline condition, and in blue are the scores for the Sonification live experiment. The points represent the mean value, whereas the line is the $90\%$ confidence interval.}
\label{fig:single_questions_live}
\end{figure}

In the case of the live experiments, $10$ people ($80\%$ M, $20\%$ F) took part in the two sessions. The users mostly come from the $20-29$ years old age group ($70\%$); with $20\%$ in the range $30-39$; and $10\%$ in $40-49$.
In this user pool, the experience with robots was rated on average around $4$ indicating a group with considerable level of experience in dealing with robots.
The live sonification experiment, whose scores are reported in Fig.~\ref{fig:single_questions_live}, validates the findings of the video-based survey. 
They confirm how the sonification of the robot's internal state is well-received and perceived as useful but the majority of the users.
Likewise, also the subjective sound qualities follow the trends observed in the previous analysis. Indeed, the sound feedback is perceived as not too interesting ($P6$) and again not very pleasant ($P7$) indicating the necessity of further tuning in the sound generation phase to achieve more user-friendly auditory feedback. 
Comparing the two live experiment conditions, Sonification and Baseline, is valuable for assessing how the robot's internal state sonification influences users' perception during the interaction.
Indeed, the mean scores to the first $3$ points, related to the robot's expressivity and communication skills, are on average $0.7$ points higher for the Sonification condition (Fig.~\ref{fig:single_questions_live} in blue) compared to the Baseline (Fig.~\ref{fig:single_questions_live} in orange).
%
\section{Conclusions}\label{sec:conclusions}
The results from both the online video-based survey and the live experiments provide precious insight into how participants perceive the expressiveness of the robot in conjunction with the auditory feedback it provides.
The video-based survey, which involved $26$ participants, revealed a general satisfaction with the robot's expressivity and behavioral understanding. Participants indicated that the auditory feedback positively influenced their interaction experience. Nonetheless, the subjective qualities of the sounds were less favorable, with many participants finding them uninteresting and unpleasant. This sentiment was echoed in additional comments, where the high-pitched sounds were described as annoying and alarming.
The live experiments, involving a smaller group of 10 participants, corroborated these findings. The sonification was again deemed useful, enhancing the robot's communication skills. However, the subjective assessment of the sounds remained consistent with the survey results, indicating a need for refinement in the auditory design to improve the user experience.
A direct comparison between the Sonification and Baseline conditions in the live experiments highlighted the benefits of sonification, with the first consistently receiving higher scores in terms of the robot's expressivity and communication skills.
In conclusion, while the sonification technique effectively enhances the user's understanding of the robot's internal state and overall interaction experience, the design of the sounds is crucial and requires improvement. 
Future work should focus on refining the auditory feedback to make it more pleasant and engaging, addressing the concerns raised by the participants, thereby ensuring more user-friendly interactions with the robot.
\section*{Acknowledgement}
This work was supported by the European Union through the project SERMAS, and by the Swiss State Secretariat for Education, Research and Innovation (SERI) under contract number 22.00247.
%
\bibliographystyle{splncs04}
\bibliography{bibliography}

\end{document}